\documentclass[11pt,letterpaper]{article}

\usepackage{newtxtext,newtxmath} % better text + math than 'times'
\usepackage{amsmath}
\usepackage{graphicx}
\usepackage{booktabs}
\usepackage{hyperref}
\usepackage[table]{xcolor}
\usepackage{float}
\usepackage{caption}
\usepackage{parskip}
\usepackage{threeparttable}
\usepackage{makecell}
\usepackage{changepage}

\title{\textbf{Learning Page Order in Shuffled WOO Releases}}

\author{
Efe Kahraman \\
e.kahraman@utf.ai
\and
Giulio Tosato \\
giulio@utf.ai
}

\date{}

\begin{document}

\maketitle

\begin{abstract}
\noindent We investigate document page ordering on 5,461 shuffled WOO documents (Dutch freedom of information releases) using page embeddings. These documents are heterogeneous collections such as emails, legal texts, and spreadsheets compiled into single PDFs, where semantic ordering signals are unreliable. We compare five methods, including pointer networks, seq2seq transformers, and specialized pairwise ranking models. The best-performing approach successfully reorders documents up to 15 pages, with Kendall's $\tau$ ranging from 0.95 for short documents (2–5 pages) to 0.72 for 15-page documents. We observe two unexpected failures: seq2seq transformers fail to generalize on long documents (Kendall's $\tau$ drops from 0.918 on 2–5 pages to 0.014 on 21–25 pages), and curriculum learning underperforms direct training by 39\% on long documents. Ablation studies suggest learned positional encodings are one contributing factor to seq2seq failure, though the degradation persists across all encoding variants, indicating multiple interacting causes. Attention pattern analysis reveals that short and long documents require fundamentally different ordering strategies, explaining why curriculum learning fails. Model specialization achieves substantial improvements on longer documents ($+0.21$ $\tau$). Code and data are publicly available on \href{https://github.com/E-8814/woo-page-ordering}{GitHub} and \href{https://huggingface.co/datasets/E-8814/woo-page-ordering}{HuggingFace} respectively.
\end{abstract}

\section{Introduction}

Dutch authorities release WOO documents (Wet open overheid) in response to public requests. A WOO document set is essentially the document trail around a topic, decision, or project, compiled from many sources. These releases are heterogeneous collections that combine emails, spreadsheets, text messages, scanned documents, and other administrative materials into a single PDF. All WOO documents within a range of 2–25 pages were collected from \href{https://open.overheid.nl}{open.overheid.nl}, yielding a dataset of 5,461 documents where the original chronological page order was known.

This study evaluates whether machine learning can recover chronological page order from content embeddings alone when pages are presented in an arbitrary order. Unlike sentence or event ordering, WOO releases are heterogeneous collages so adjacent pages often lack semantic continuity. Per-page metadata (e.g., timestamps, page numbers, thread identifiers) is frequently missing or unreliable. Many pages contain no dates, and when dates do appear they are often ambiguous or refer to quoted history (emails, legal citations) rather than the page's true position.

Each page was embedded using text-embedding-3-large, and ordering quality was evaluated using Kendall's tau. The number of possible page orderings grows factorially with document length: 2 pages have 2 possible orderings, while 25 pages have $25! \approx 1.55 \times 10^{25}$ possible orderings.

Five methods were evaluated across eleven model configurations: geometric heuristics, neural position prediction, pointer networks, seq2seq transformers, and pairwise ranking transformers, including length-specialized variants optimized for specific document ranges. We also compared curriculum learning against direct training.

\section{Related Work}

The problem of ordering collaged PDFs is closely related to page-stream segmentation (PSS), which detects document boundaries in continuous page streams \cite{gordo2013,heusden2024openpss,braz2021}. While PSS focuses on splitting a stream into separate documents, we study recovering the page order within a given set of documents. 

Existing reading-order models are designed to determine the sequence of text segments inside a single document, such as
navigating complex newspaper layouts or forms \cite{wang2021layoutreader,zhang2024roor,peng2022ernielayout}, but they do not address cross-page sequencing, where cues are typically noisier.

Prior work on sentence, event, and story ordering assumes that consecutive elements have semantic connections and follow logical time sequences \cite{barzilay2008,li2017,chambers2008,ning2017,mostafazadeh2016}. For example, sentences in a story naturally flow together, and events happen in cause-and-effect order. Recent methods use pairwise ranking models \cite{kumar2020} and constraint graphs \cite{zhu2021} to capture these ordering relationships. WOO releases are fundamentally different: pages come from mixed document types that don't naturally connect to each other, making standard semantic ordering signals much weaker.

From a modeling perspective, this is a permutation learning problem. Pointer networks \cite{vinyals2015} generate orderings autoregressively by selecting one element at a time from the remaining candidates, while Transformers \cite{vaswani2017} have been applied to combinatorial permutation tasks such as routing \cite{bello2016,kool2019}. An alternative is pairwise ranking, where the model predicts ``comes-before'' relations for every pair and aggregates them into a global ordering \cite{dwork2001}. This decomposition reduces the problem from predicting a full permutation to making independent binary comparisons, which scales more naturally to longer sequences.

Positional encoding affects length generalization. Transformers treat sequence length as a domain: models trained on short sequences develop length-specific biases that cause performance to collapse on longer inputs, even when the underlying task remains the same \cite{varis2021}. A systematic comparison of five positional encoding schemes found that none consistently extrapolates well, with even no positional encoding sometimes outperforming explicit schemes \cite{kazemnejad2023}. 

Curriculum learning presents training examples in a structured progression from simple to complex cases, aiming to stabilize learning and improve generalization \cite{elman1993,krueger2009,bengio2009}. However, curriculum benefits are task-dependent and can even be counterproductive \cite{wu2021curricula,hacohen2019}: when the skills learned from simple examples differ fundamentally from those needed for complex cases, the model develops strategies that fail to transfer.

\section{Experimental Setup}

\subsection{Dataset}

The dataset consists of 5,461 WOO documents with the following length distribution:

\begin{table}[H]
\begin{adjustwidth}{-1.2cm}{-1.2cm}
\captionsetup{justification=centering}
\centering
\caption{Dataset length distribution}
\label{tab:dataset}
\small
\rowcolors{2}{gray!10}{white}
\begin{tabular}{@{}lrr@{}}
\toprule
\rowcolor{white}
Length Range & Document Count & Percentage \\
\midrule
2--5 pages   & 1,247 & 22.8\% \\
6--10 pages  & 1,683 & 30.8\% \\
11--15 pages & 1,204 & 22.0\% \\
16--20 pages & 789   & 14.4\% \\
21--25 pages & 538   & 9.9\%  \\
\midrule
\rowcolor{white}
\textbf{Total} & \textbf{5,461} & \textbf{100\%} \\
\bottomrule
\end{tabular}
\end{adjustwidth}
\end{table}

This distribution reflects the complete population of WOO documents in the 2--25 page range available on \href{https://open.overheid.nl}{open.overheid.nl}, not a sample. The skew toward shorter documents is an inherent property of the domain.

Text was extracted from PDFs using PyMuPDF with OCR fallback (Tesseract, Dutch, 800 DPI), then embedded using OpenAI's text-embedding-3-large (3072 dims). Only textual content was embedded, as the embedding model only supports text; visual elements such as charts, graphs, and tables were excluded. Due to the dataset scale, manual segmentation was impractical; thus, each page is treated as an independent unit, even though documents often contain multi-page logical units that ideally would be embedded together. The dataset is split into: train (70\%), validation (15\%), and test (15\%).

\subsection{Evaluation}

Kendall's tau ($\tau$) \cite{kendall1938} measures rank correlation between predicted and ground truth orderings:
\[\tau = \frac{\text{concordant pairs} - \text{discordant pairs}}{\text{total pairs}}\]

A pair of pages $(i,j)$ is concordant if the model predicts the same relative order as the ground truth (both place $i$ before $j$), and discordant if the relative order is reversed. Kendall's tau $\tau$ ranges from $-1$ (perfect reversal) to $+1$ (perfect agreement). It rewards partially-correct orderings and evaluates relative order rather than exact absolute positions.

\subsection{Methods Compared}

\textbf{1. Heuristics:}\label{method:1} Random baseline, greedy nearest neighbor (starts from a random page and repeatedly appends the most similar unvisited page in embedding space), TSP nearest neighbor (treats ordering as a traveling salesman problem using nearest neighbor heuristic).

\textbf{2. BiLSTM Position Classifier:}\label{method:2} A bidirectional LSTM \cite{hochreiter1997} that processes all page embeddings and independently predicts a position score for each page (e.g., ``this page looks like position 3''). Pages are then sorted by their predicted scores to produce the final ordering. Unlike sequential methods, all predictions are made simultaneously without considering previously selected pages.

\textbf{3. Pointer Networks:}\label{method:3} Pointer networks \cite{vinyals2015} build an ordering by picking one page at a time from the remaining candidates. At each step, the model looks at every page that has not yet been placed, assigns each one a score indicating how likely it is to come next, and selects the highest-scoring page. That page is then removed from the candidate pool, and the process repeats until all pages have been placed. Because the output is always a selection from the input rather than a generated token, the result is guaranteed to be a valid permutation. We implemented two variants that differ in how pages are represented and how the model decides what comes next:
\begin{itemize}
\item[\textbf{3.1}]\label{method:3.1} \textbf{Pointer MLP:} A simplified variant that uses only feedforward layers (3.1M parameters). Each page embedding is independently transformed through a three-layer network to produce a richer representation. At each step, the model compares every remaining page against a decoder state, picks the best match, and then updates the decoder state based on the page just selected. Because there is no recurrent memory, each decision is based only on the most recently chosen page---the model does not remember earlier selections.
\item[\textbf{3.2}]\label{method:3.2} \textbf{Pointer LSTM:} The classic pointer network architecture \cite{vinyals2015} (9.5M parameters). A bidirectional LSTM encoder first reads all page embeddings together, so each page's representation is informed by the pages around it. An LSTM decoder then selects pages one by one, carrying a hidden state that accumulates information from every previous selection. This means each decision considers the full history of what has already been placed, not just the most recent choice.
\end{itemize}

\textbf{4. seq2seq Transformer:}\label{method:4} This approach uses a sequence to sequence transformer \cite{vaswani2017} to map shuffled pages to an ordered sequence. The encoder uses self-attention to build representations of all pages simultaneously, while the decoder generates the ordering one position at a time, attending to the encoder output to decide which page comes next. We implemented three variants to study the effect of positional encodings:
\begin{itemize}
\item[\textbf{4.1}]\label{method:4.1} \textbf{Learned positional encodings:} Before processing, the model adds a position signal to each page so it can tell apart ``the page in slot 1'' from ``the page in slot 2,'' and so on. These signals start out random and are gradually refined during training. The problem is that the model can only learn useful signals for positions it sees often. Position 1 appears in every document, but position 24 only appears in the longest ones, so the model never learns a reliable signal for high positions.
\item[\textbf{4.2}]\label{method:4.2} \textbf{Sinusoidal positional encodings:} Instead of learning position signals from data, this variant uses fixed mathematical wave patterns \cite{vaswani2017} that produce a unique signal for every position regardless of whether it was seen during training. Because these patterns are predefined rather than learned, they allow better generalization to document lengths that are underrepresented in the training data.
\item[\textbf{4.3}]\label{method:4.3} \textbf{No positional encodings:} Removes positional encodings entirely, forcing the model to rely solely on page content without any position information. This ablation tests whether positional encoding is responsible for the transformer's failure on long documents.
\end{itemize}

\textbf{5. Pairwise Ranking Transformer:}\label{method:5} Instead of predicting the full sequence directly, this approach answers a simpler question for every pair of pages: ``Should page $j$ come after page $i$?'' First, all page embeddings pass through transformer encoder layers with multi-head self-attention, allowing each page representation to incorporate context from all other pages. Then, for each ordered pair $(i,j)$, the model computes the difference between their contextualized embeddings (${d}_{ij} = \text{encoded}_j - \text{encoded}_i$) and feeds it through a 4-layer scoring network that outputs a single score $s_{ij}$ representing how strongly $j$ should follow $i$. This produces an $N \times N$ matrix of pairwise ``comes-after'' scores. To convert these pairwise judgments into a final ordering, each page receives an aggregated position score:
\[
\text{score}_i = \frac{1}{N}\sum_j s_{ji} - \frac{1}{N}\sum_j s_{ij}
\]
which captures how many pages should precede $i$ (first term) minus how many should follow $i$ (second term). Pages are then sorted by descending position scores. This non-autoregressive design makes all pairwise predictions simultaneously, avoiding the error accumulation that can occur when generating sequences one step at a time. We implemented three variants to compare training strategies:
\begin{itemize}
\item[\textbf{5.1}]\label{method:5.1} \textbf{Universal model:} Trained on documents of all lengths with uniform weighting (531M parameters).
\item[\textbf{5.2}]\label{method:5.2} \textbf{Specialized models (direct training):} Five separate pairwise ranking transformers, each with architecture capacity scaled to its target document length. Longer documents require deeper networks to capture longer-range dependencies and wider hidden dimensions to maintain distinct representations across more pages. The specialized models scale architecture capacity with target length (layers / hidden dimension / parameters): 2--5 pages (6 / 1024 / 80.5M), 6--10 (8 / 1536 / 235M), 11--15 (10 / 2048 / 517M), 16--20 (12 / 2048 / 581M), 21--25 (12 / 3072 / 1190M), totaling $\sim$2.6B parameters. Each model trains on all document lengths (2--25 pages) but applies $5\times$ loss weighting on its target range. This weighting balances two goals: the model still learns from all document lengths (preventing overfitting to a single range), while being strongly guided toward its specialization target. At inference, documents are routed to their length-appropriate model.
\item[\textbf{5.3}]\label{method:5.3} \textbf{Specialized models (curriculum learning):} Same five-model ensemble with identical architectures and parameter counts. Instead of direct training, each model follows a multi-stage curriculum that progressively increases document length from short examples to its target range. Training starts on short documents (2--5 pages), gradually expands the length range across multiple stages, then focuses exclusively on the target range for the remaining epochs with a reduced learning rate. For example, the 6--10 page model progresses through four stages: 2--5 pages, 4--7 pages, and then 6--10 pages for the majority of training.
\end{itemize}

\section{Results}

\begin{figure}[H]
\begin{adjustwidth}{-1.2cm}{-1.2cm}
\centering
\includegraphics[width=\linewidth]{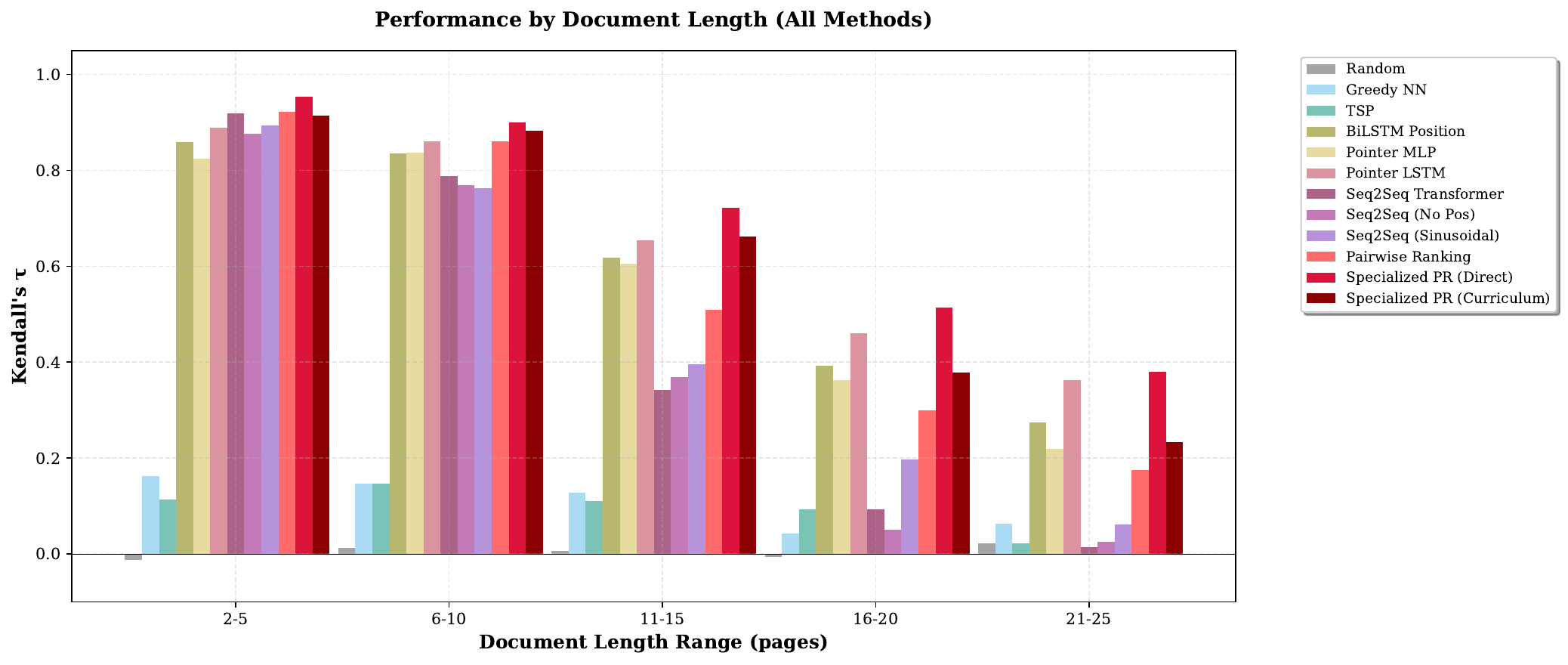}
\end{adjustwidth}
\caption{Kendall's $\tau$ by method and document length range.}
\label{fig:grouped_bars}
\end{figure}

\begin{figure}[H]
\begin{adjustwidth}{-1.2cm}{-1.2cm}
\centering
\includegraphics[width=\linewidth]{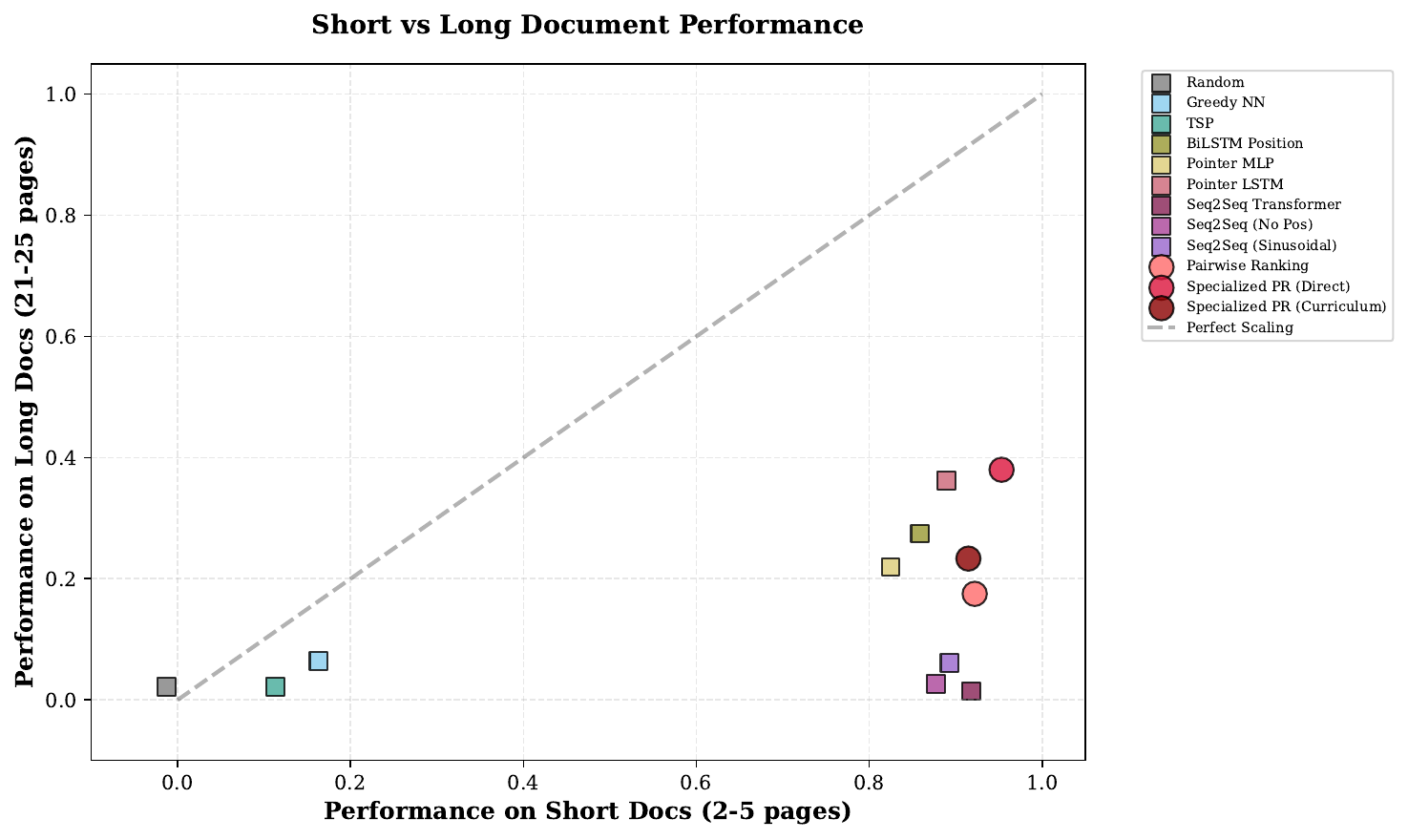}
\end{adjustwidth}
\caption{Short document (2--5 pages) vs long document (21--25 pages) performance. Points below the diagonal indicate failure to scale.}
\label{fig:short_vs_long}
\end{figure}

Table~\ref{tab:results} shows Kendall's tau by document length on the test set, with performance trends visualized in Figure~\ref{fig:grouped_bars} and short-versus-long scaling behavior in Figure~\ref{fig:short_vs_long}.

\vspace{43pt}

\begin{table}[H]
\begin{adjustwidth}{0 cm}{0 cm}
\captionsetup{justification=centering}
\centering
\begin{threeparttable}
\caption{Kendall's tau by document length (test set)}
\label{tab:results}
\small
\rowcolors{2}{gray!10}{white}
\begin{tabular}{@{}lrrrrrc@{}}
\toprule
\rowcolor{white}
Model & \makecell{2--5\\pages} & \makecell{6--10\\pages} & \makecell{11--15\\pages} & \makecell{16--20\\pages} & \makecell{21--25\\pages} & Params \\
\midrule
Random & 0.007 & 0.002 & $-0.001$ & 0.003 & 0.001 & 0 \\
Greedy NN & 0.168 & 0.091 & 0.062 & 0.045 & 0.033 & 0 \\
TSP NN & 0.113 & 0.147 & 0.111 & 0.093 & 0.022 & 0 \\
BiLSTM Position & 0.859 & 0.667 & 0.503 & 0.402 & 0.318 & 3.7M \\
Pointer MLP & 0.847 & 0.682 & 0.551 & 0.448 & 0.371 & 3.1M \\
Pointer LSTM & 0.889 & 0.703 & 0.572 & 0.461 & 0.362 & 9.5M \\
seq2seq (learned) & 0.918 & 0.787 & 0.343 & 0.094 & 0.014 & 45M \\
seq2seq (sinusoidal) & 0.893 & 0.763 & 0.396 & 0.197 & 0.061 & 45M \\
seq2seq (no position) & 0.877 & 0.770 & 0.369 & 0.051 & 0.026 & 45M \\
Pairwise Ranking & 0.922 & 0.860 & 0.509 & 0.300 & 0.175 & 531M \\
Specialized PR (Direct) & 0.953 & 0.899 & 0.722 & 0.515 & 0.380 & $\sim$2.6B \\
Specialized PR (Curriculum) & 0.915 & 0.882 & 0.662 & 0.379 & 0.233 & $\sim$2.6B \\
\bottomrule
\end{tabular}
\end{threeparttable}
\end{adjustwidth}
\vspace{8pt}
{\small\noindent Note: Parameters for Specialized PR are summed across the five-model ensemble. Each document is routed to one specialized model at inference.}
\end{table}

\vspace{20pt}

\subsection{Performance Overview}

The specialized pairwise ranking transformer (\hyperref[method:5.2]{method 5.2}) achieves strong performance on documents up to 15 pages, with $\tau = 0.953$ on 2--5 pages, $\tau = 0.899$ on 6--10 pages, and $\tau = 0.722$ on 11--15 pages (Table~\ref{tab:results}). This shows consistent improvements over autoregressive approaches. The pairwise ranking architecture outperforms pointer networks (\hyperref[method:3]{method 3}) by $+0.064$ $\tau$ on short documents and $+0.150$ $\tau$ on 11--15 page documents. This supports the advantage of non-autoregressive pairwise prediction over sequential generation. However, performance degrades on longer documents (16--25 pages), where even the specialized model achieve only moderate accuracy.

Methods exhibit distinct scaling behaviors across document lengths. The seq2seq transformer shows catastrophic degradation: $\tau = 0.918$ on short documents drops to $\tau = 0.014$ on 21--25 pages. Curriculum learning consistently underperforms direct training across all ranges. In contrast, specialized training maintains relatively consistent performance as document length increases, while pointer LSTM demonstrates graceful degradation.

\subsection{Limits of Heuristics}

Greedy nearest neighbor, TSP nearest neighbor (\hyperref[method:1]{method 1}) achieve $\tau < 0.17$ at all lengths (Figure~\ref{fig:short_vs_long}), indicating that adjacent pages in correct order are not close in embedding space. This occurs because WOO documents contain mixed content types where page boundaries occur at arbitrary content breaks rather than semantic transitions. A legal document's page 5 might be closer in embedding space to an unrelated email than to its own page 6.

\subsection{The seq2seq Transformer Failure}

\begin{figure}[H]
\begin{adjustwidth}{-1.2cm}{-1.2cm}
\centering
\includegraphics[width=\linewidth]{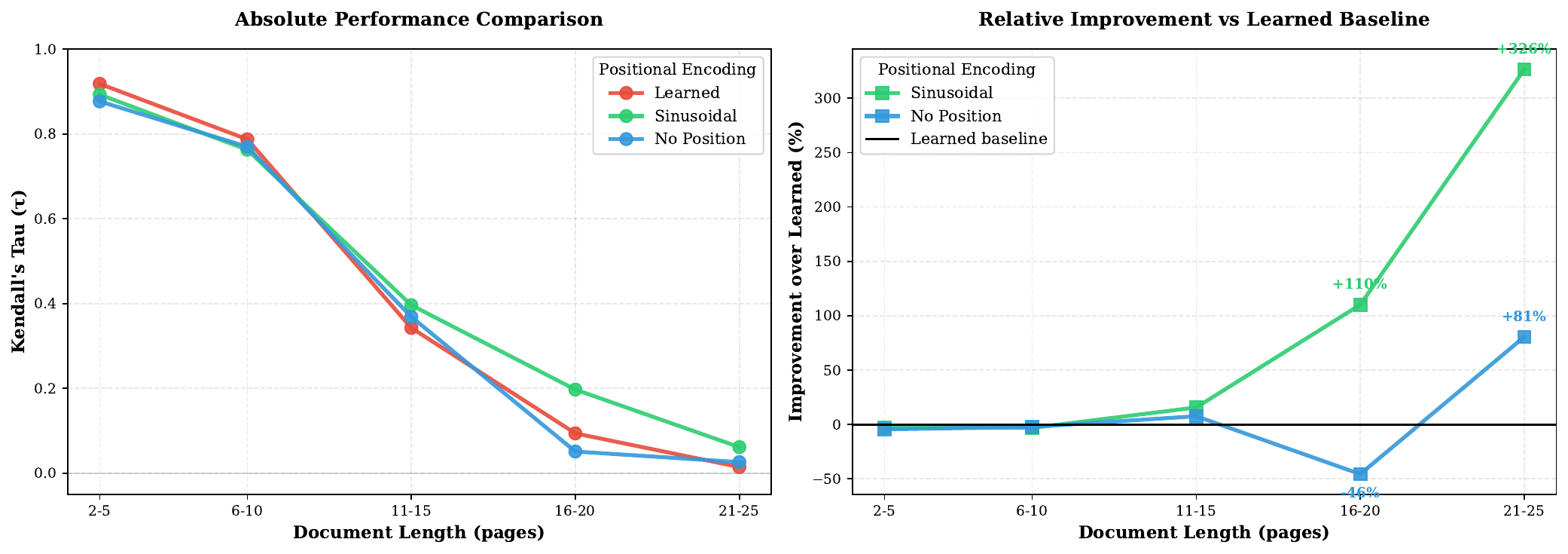}
\end{adjustwidth}
\caption{seq2seq transformer positional encoding ablations. Left: Absolute Kendall's $\tau$ by document length. Right: Relative improvement over learned baseline. Notably, no positional encodings helps on medium-length documents but fails on long documents.}
\label{fig:seq2seq_ablation}
\end{figure}

The seq2seq transformer (\hyperref[method:4]{method 4}) exhibits severe performance degradation on long documents ($\tau = 0.014$ on 21--25 pages versus $0.918$ on 2--5 pages). To better understand this failure, we conducted ablation studies on positional encodings, training three variants: learned embeddings (\hyperref[method:4.1]{4.1}), sinusoidal encodings (\hyperref[method:4.2]{4.2}), and no positional encodings (\hyperref[method:4.3]{4.3}). Learned positional encodings are trained only on positions that appear in the training data; position 24 appears in about $\tfrac{1}{45}$ as many documents as position 0, causing the model to poorly represent rarely-seen positions. Sinusoidal encodings moderately improve long-document performance ($\tau$ increases from 0.014 to 0.061 on 21--25 pages, a 326\% improvement, though both values remain near zero; Figure~\ref{fig:seq2seq_ablation}), suggesting that learned positional encodings struggle to extrapolate beyond the training distribution. However, the model still fails on long documents even with sinusoidal encodings, indicating the problem is multi-causal rather than stemming from positional encodings alone.

\begin{figure}[H]
\begin{adjustwidth}{-1.2cm}{-1.2cm}
\centering
\includegraphics[width=\linewidth]{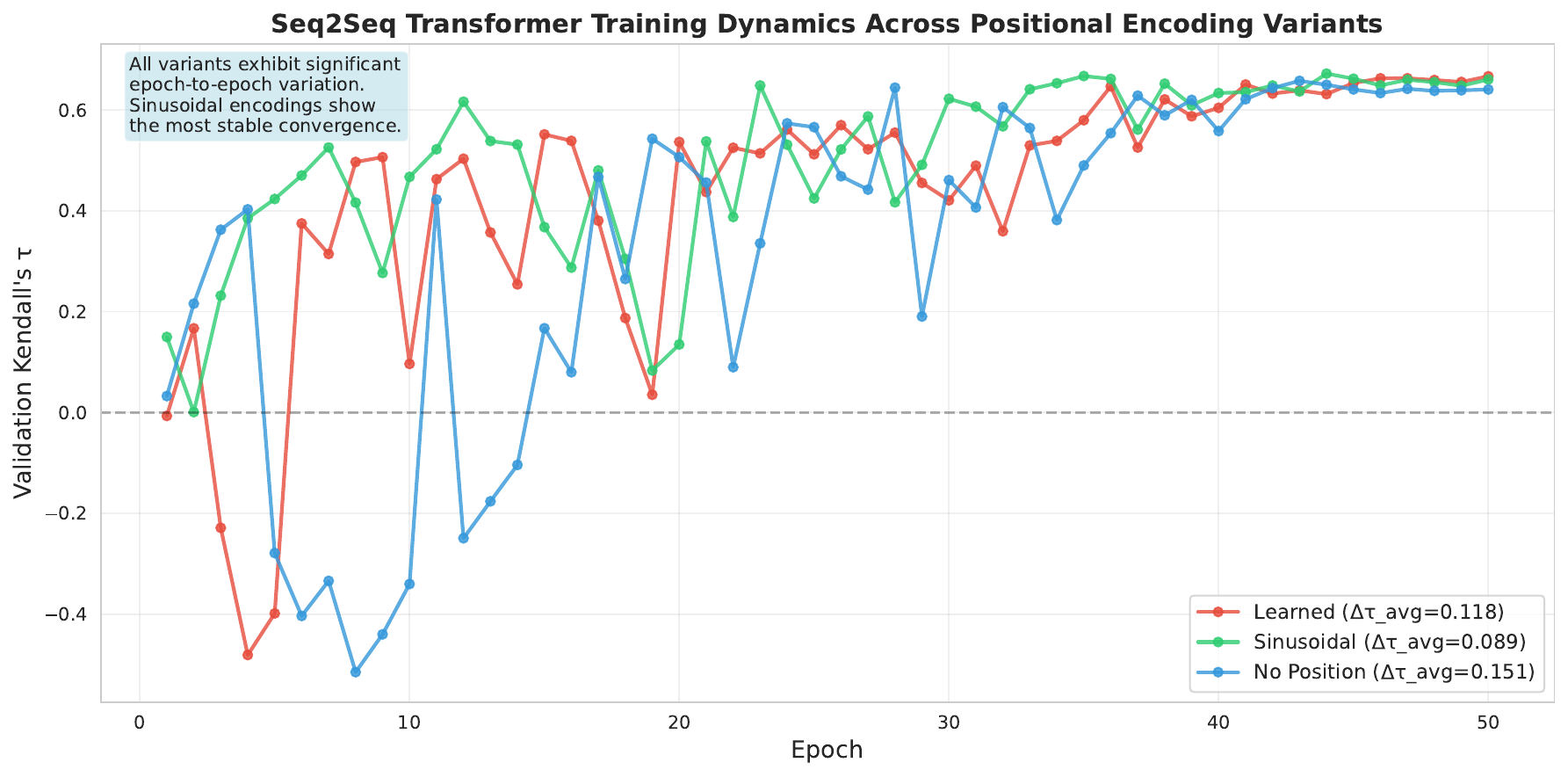}
\end{adjustwidth}
\caption{Training dynamics for seq2seq positional encoding variants. Validation Kendall's $\tau$ across epochs shows high oscillation for all variants, with sinusoidal encodings providing the most stable training.}
\label{fig:training_instability}
\end{figure}

Training dynamics show considerable instability across all variants (Figure~\ref{fig:training_instability}), with validation performance oscillating significantly across consecutive epochs (standard deviation $\sigma = 0.262$ for learned encodings, improving to $\sigma = 0.169$ with sinusoidal encodings). Removing positional encodings entirely worsens training stability ($\sigma = 0.347$), suggesting positional information provides necessary training signal despite imperfect extrapolation. Sinusoidal encodings never reach negative validation performance, while learned and no-position variants occasionally produce $\tau < 0$, indicating negative $\tau$ (worse-than-random) predictions.

We hypothesize that this failure arises from a combination of training data imbalance, positional encoding limitations, and architectural depth. Among these, positional encodings contribute substantially to the failure, but do not fully explain it.

This contrasts with pointer networks (\hyperref[method:3]{method 3}), which use autoregressive decoding but exhibit a more gradual degradation in performance. Pointer networks employ shallower architectures (2--3 layers, $<$10M params) with simple single-layer attention, while the seq2seq transformer uses deep encoder-decoder stacks (6 layers each) with multi-head self-attention and cross-attention. The architectural differences, combined with positional encoding choices, may explain the divergent scaling behavior.

\subsection{Model Specialization Benefits}

Training separate models for specific document length ranges (\hyperref[method:5.2]{method 5.2}) substantially improves performance over the universal pairwise ranking baseline (\hyperref[method:5.1]{method 5.1}). The specialized ensemble uses the same architecture but trains five models with $5\times$ weighted loss on their target ranges.

Performance gains increase with document length: $+0.031$ $\tau$ on 2--5 pages, growing to $+0.205$ $\tau$ on 21--25 pages. This pattern suggests that longer documents benefit more from specialized optimization, possibly because they require different representational strategies than shorter documents. The universal model must learn a single strategy that works across all lengths, which may compromise performance on the underrepresented longer documents.

The $5\times$ weighting allows each model to see all document lengths (preventing overfitting to a single length) while emphasizing its target range. This appears more effective than the universal model's uniform weighting, particularly for underrepresented long documents. The specialized models achieve $\tau = 0.380$ on 21--25 pages, compared to $\tau = 0.175$ for the universal baseline, a $2.2\times$ improvement.

\subsection{Why Does Curriculum Learning Fail?}

Curriculum learning theory \cite{bengio2009} suggests that learning simple examples first facilitates learning on complex examples. Yet curriculum learning (\hyperref[method:5.3]{method 5.3}) consistently underperforms direct training (\hyperref[method:5.2]{method 5.2}), with the performance gap widening on longer documents where curriculum should help most (39\% worse on 21--25 pages).

To investigate this failure, we conducted two validation experiments. First, we analyzed attention patterns in specialized models trained on short (2--5 pages) versus long (21--25 pages) documents. Short models exhibit highly local attention (77.9\% of attention within $\pm$2 positions, average distance 1.53), while long models attend globally (20.8\% local attention, average distance 7.59), representing a $4.96\times$ difference. Second, we trained a model exclusively on short documents ($\tau = 0.8817$) and tested it on long documents, achieving $\tau = 0.1618$, demonstrating weak transfer. These results suggest that short and long documents require different ordering strategies, which explains why curriculum learning, by forcing the model to learn the local strategy first, underperforms direct training by 39\% on long documents.

\section{Limitations and Future Work}

Several aspects of our approach require further investigation. Our embedding pipeline processes only textual content, excluding visual elements such as charts, graphs, and tables that may contain important ordering signals. We run all experiments using only one embedding model (text-embedding-3-large). Results might be highly dependent on the performance of the embedding model. Pages are treated as independent units, though many WOO documents contain multi-page logical structures that would benefit from joint embedding. Furthermore, while we identify learned positional encodings as a contributing factor, the complete explanation for the seq2seq transformer's catastrophic failure on long documents remains unclear. The natural skew toward shorter documents ($2.3\times$ more short than long documents) confounds performance comparisons across length ranges, as lower performance on longer documents may partially reflect data scarcity rather than task difficulty alone.

Future work could incorporate multimodal embeddings to capture both text and visual features, develop methods for automatic segmentation of multi-page logical units, and investigate alternative transformer architectures with better length extrapolation properties such as ALiBi \cite{press2022} or RoPE \cite{su2021}.

\section{Conclusion}

We compared eleven model configurations on 5,461 WOO documents. The specialized pairwise ranking transformer achieved strong results on documents up to 15 pages ($\tau = 0.953$ on 2--5 pages, $\tau = 0.722$ on 11--15 pages), outperforming pointer networks by up to $+0.150$ $\tau$. Seq2seq transformers fail to generalize on long documents ($\tau = 0.918 \rightarrow \tau = 0.014$); ablation studies implicate learned positional encodings as one factor, but the failure persists across encoding variants, pointing to deeper architectural limitations. Curriculum learning underperformed direct training because short and long documents require incompatible strategies (local versus global attention), preventing transfer. Specialized training improved performance substantially ($+0.21$ $\tau$ on long documents). 

These findings show that document ordering on heterogeneous collections presents unique challenges, and that architectural choices and training strategies significantly impact generalization to longer sequences.

\pagebreak

\end{document}